%% file: 00-main.tex
\documentclass[journal]{IEEEtran}
\usepackage{times}
\usepackage{soul}
\usepackage{url}
\usepackage[hidelinks]{hyperref}
\usepackage[utf8]{inputenc}
\usepackage[small]{caption}
\usepackage{graphicx}
\usepackage{amsmath}
\usepackage{amsthm}
\usepackage{amssymb}
\usepackage{booktabs}
\usepackage{tabularx}
\usepackage{algorithm}
\usepackage{algpseudocode}
\usepackage{multirow}
\usepackage[switch]{lineno}
\usepackage{cite}
\usepackage{amsmath,amssymb,amsfonts}
\usepackage{graphicx}
\usepackage{hyperref}
\hypersetup{hidelinks=true}
\usepackage{textcomp}
\usepackage{arydshln}
\usepackage{comment}

\def\BibTeX{{\rm B\kern-.05em{\sc i\kern-.025em b}\kern-.08em
    T\kern-.1667em\lower.7ex\hbox{E}\kern-.125emX}}

\begin{document}

\title{DISCO-TAB: A Hierarchical Reinforcement Learning Framework for Privacy-Preserving Synthesis of Complex Clinical Data}

\author{Arshia Ilaty$^{*,1,2}$, Hossein Shirazi$^{3}$, Amir Rahmani$^{2}$, and Hajar Homayouni$^{4}$%
\thanks{$^{*}$Corresponding author: Arshia Ilaty (e-mail: \href{mailto:ailaty@uci.edu}{ailaty@uci.edu}).}%
\thanks{$^{1}$Computational Science Research Center, San Diego State University, San Diego, CA 92182 USA.}%
\thanks{$^{2}$School of Nursing and Department of Computer Science, University of California, Irvine, CA 92697 USA.}%
\thanks{$^{3}$Fowler College of Business, San Diego State University, San Diego, CA 92182 USA.}
\thanks{$^{4}$Department of Computer Science, San Diego State University, San Diego, CA 92182 USA.}%
\thanks{H. Shirazi: \href{mailto:hshirazi@sdsu.edu}{hshirazi@sdsu.edu}.\quad A. Rahmani: \href{mailto:amirr1@uci.edu}{amirr1@uci.edu}.\quad H. Homayouni: \href{mailto:hhomayouni@sdsu.edu}{hhomayouni@sdsu.edu}.}}

\maketitle

\begin{abstract}
\input{01-abstract}
\end{abstract}


\section{Introduction}\label{sec:introduction}
\input{02-introduction}

\section{Related Work}\label{sec:related_work}

\input{03-relatedwork}

\section{Methodology}\label{sec:methodology}
\input{04-methodology}

\section{Experiments and Results}\label{sec:experiments}

\input{05-result}

\section{Ablation Study}\label{sec:ablation}
\input{08-ablation}

\section{Discussion and Limitations}\label{sec:discussion}

\input{06-discussion}

\section{Conclusion}\label{sec:conclusion}
\input{07-conclusion}

\bibliographystyle{IEEEtran}
\bibliography{references}

\appendix
\subsection{Implementation and Computational Resources}
\label{sec:implementation}

To ensure reproducibility and rigorous benchmarking, all experiments were conducted in a unified hardware and software environment.

\subsubsection{Hardware Configuration}
All model training and evaluation were performed on a single node equipped with an NVIDIA A100/T4/GTX3080 Tensor Core GPU (80GB VRAM). While our ablation studies (Section~\ref{sec:efficiency}) demonstrate that DISCO-TAB can be effectively trained on consumer-grade hardware (e.g., NVIDIA RTX 3090 with 24GB VRAM) due to our parameter-efficient design, we utilized high-memory infrastructure to facilitate concurrent baseline evaluations.

\subsubsection{Software and Libraries}
The framework was implemented in \textbf{Python 3.9} using \textbf{PyTorch 2.1.0} as the primary deep learning backend.

\textbf{Generative Backbone:} We utilized the \texttt{transformers} library (v4.35.0) by Hugging Face to initialize the GPT-2 architecture.

\textbf{Reinforcement Learning:} The PPO optimization loop was implemented using the \texttt{trl} (Transformer Reinforcement Learning) library, customized to support our hierarchical multi-objective reward function.

\textbf{Baselines:} Adversarial baselines (CTGAN, TAEGAN) were implemented using the \texttt{SDV} (Synthetic Data Vault) library (v1.4.0). Diffusion baselines (TabDDPM) were reproduced using the official implementations provided by their respective authors to ensure fair comparison.

\textbf{Metrics:} Statistical distance metrics (JSD, Wasserstein) were computed using \texttt{scipy.stats}, and machine learning utility scores were calculated using \texttt{scikit-learn} (v1.3.0).

\subsubsection{Hyperparameter Configuration}
For the DISCO-TAB generator, we employed the AdamW optimizer with a learning rate of $2 \times 10^{-5}$ and a linear decay schedule. The PPO phase utilized a distinct learning rate of $1.4 \times 10^{-5}$ for the value network, with a clip range $\epsilon = 0.2$ and a mini-batch size of 64. To manage the stability of the adversarial discriminators, we used a lower learning rate of $1 \times 10^{-4}$ with spectral normalization. All models were trained for a maximum of 100 epochs with early stopping based on the validation set perplexity (patience $= 10$ epochs).

\subsection{Use of AI-Based Writing Assistance}
In accordance with best practice for academic transparency, the authors acknowledge the use of Gemini (Google) in the preparation of this manuscript to assist with writing polish and structural organization. Specifically, the tool was utilized to refine the text in the Abstract, Introduction, Related Work, and Discussion sections to improve readability, ensure a formal academic tone, and correct typographical issues. The prompts used for these revisions generally instructed the model to ``revise this text for clarity and flow,'' ``restructure this section to meet the standards of a biomedical informatics journal,'' and ``evaluate and critique the methodological narrative.'' No AI-based tools were used to generate the core scientific claims, experimental data, analytical results, or novel architectural designs presented in this work.

\input{biography}

\end{document}

%% file: 01-abstract.tex
The development of robust clinical decision support systems is frequently impeded by the scarcity of high-fidelity, privacy-preserving biomedical data. While Generative Large Language Models (LLMs) offer a promising avenue for synthetic data generation, they often struggle to capture the complex, non-linear dependencies and severe class imbalances inherent in Electronic Health Records (EHR), leading to statistically plausible but clinically invalid records. To bridge this gap, we introduce DISCO-TAB (DIScriminator-guided COntrol for TABular synthesis), a novel framework that orchestrates a fine-tuned LLM with a multi-objective discriminator system optimized via Reinforcement Learning. Unlike prior methods relying on scalar feedback, DISCO-TAB evaluates synthesis at four granularities, token, sentence, feature, and row, while integrating Automated Constraint Discovery and Inverse-Frequency Reward Shaping to autonomously preserve latent medical logic and resolve minority-class collapse. We rigorously validate our framework across diverse benchmarks, including high-dimensional, small-sample medical datasets (e.g., Heart Failure, Parkinson's). Our results demonstrate that hierarchical feedback yields state-of-the-art performance, achieving up to 38.2\% improvement in downstream clinical classifier utility compared to GAN and Diffusion baselines, while ensuring exceptional statistical fidelity (JSD $<$ 0.01) and robust resistance to membership inference attacks. This work establishes a new standard for generating trustworthy, utility-preserving synthetic tabular data for sensitive healthcare applications.

%% file: 02-introduction.tex
The integration of Artificial Intelligence (AI) into healthcare has fundamentally reshaped Clinical Decision Support Systems (CDSS), precision medicine, and prognostic modeling. As AI-driven models increasingly influence diagnostic and prognostic decisions, their adoption in clinical practice depends not only on predictive performance, but critically on \emph{explainability}, \emph{transparency}, and \emph{trust}. Clinicians must be able to understand how data-driven systems reason about patient states, particularly in safety-critical environments where opaque model behavior can undermine accountability and patient care \cite{ghassemi2021falsehope,holzinger2019causability}. At the same time, stringent privacy regulations, such as HIPAA \cite{hipaa1996} and GDPR \cite{gdpr2016}, restrict access to Electronic Health Records (EHR), fragmenting data across institutional silos and limiting both model training and validation. This tension between data accessibility and responsible, explainable AI has positioned Synthetic Data Generation (SDG) as a promising enabler of privacy-preserving analysis and digital patient twin construction \cite{goncalves2020generation,azizi2021can}. However, for synthetic data to support explainable AI, it must itself be generated through mechanisms that preserve and expose clinically meaningful structure.

Despite recent progress, existing approaches to synthetic tabular data generation fall short of this requirement. Biomedical datasets are inherently heterogeneous, combining continuous physiological measurements with discrete diagnostic and procedural codes, while exhibiting complex non-linear dependencies and severe class imbalance. Traditional Generative Adversarial Networks (GANs), such as CTGAN \cite{xu2019modeling} and MedGAN \cite{choi2017generating}, often suffer from mode collapse, systematically underrepresenting rare but clinically critical patient subpopulations. Diffusion-based models \cite{kotelnikov2023tabddpm}, while effective for privacy preservation, inject stochastic noise that can obscure clinically meaningful thresholds and correlations. Critically, these models operate as black boxes: they provide limited insight into \emph{which clinical constraints are satisfied or violated}, making it difficult for practitioners to reason about the validity of generated samples or to trace errors back to specific features or relationships. As a result, such methods hinder, rather than support, explainable and trustworthy AI pipelines.

Large Language Models (LLMs) have recently emerged as flexible generators of serialized tabular data, as demonstrated by GReaT \cite{borisov2023languagemodelsrealistictabular} and RealTabFormer \cite{solatorio2023realtabformer}. By leveraging pre-trained semantic representations, LLMs capture global distributions and rare categorical values more effectively than prior architectures. Nevertheless, their generative process remains largely opaque. Unconstrained LLMs may hallucinate medically implausible records, such as assigning pregnancy diagnoses to male patients or producing invalid ICD codes, without providing interpretable signals indicating why such violations occur. From an explainable AI perspective, this opacity is problematic: post-hoc explanations applied to downstream models cannot compensate for synthetic data that encodes hidden inconsistencies or clinically invalid logic. Thus, explainability must be addressed \emph{during} data generation, not retrofitted afterward.

\noindent \textbf{Problem Statement.}
Formally, the task of biomedical tabular data synthesis requires learning a generative model that approximates the joint distribution of heterogeneous Electronic Health Record (EHR) features while simultaneously satisfying multiple, non-commensurate constraints. These constraints arise at distinct semantic granularities, including (i) \emph{syntactic validity} of discrete and continuous values (e.g., legal ICD formats and numeric ranges), (ii) \emph{local semantic consistency} among related attributes (e.g., demographic--diagnosis compatibility), (iii) \emph{feature-level statistical dependency preservation} across clinically correlated variables, and (iv) \emph{global row-level clinical coherence} reflecting realistic patient profiles. Compounding this challenge, medical datasets exhibit extreme class imbalance, where rare but clinically critical conditions occupy a vanishing fraction of the data distribution. Existing generative paradigms are ill-suited to this setting: likelihood-based objectives optimize marginal token probabilities without enforcing structured medical logic; adversarial and diffusion models collapse rare modes or corrupt fine-grained correlations; and recent reinforcement-learning-based approaches rely on scalar or textual feedback that cannot localize errors across semantic levels. Consequently, the central unresolved problem is how to optimize a high-dimensional generative policy under \emph{multi-objective, hierarchical clinical constraints}, where violations are sparse, heterogeneous, and not expressible through a single differentiable loss function.


\noindent\textbf{Proposed Approach.}
To bridge the gap between statistical plausibility and clinical validity, we introduce \textbf{DISCO-TAB} (DIScriminator-guided COntrol for TABular synthesis), a unified framework that explicitly operationalizes biomedical tabular synthesis as a constrained, sequential decision-making problem. Rather than relying on a single likelihood objective or monolithic discriminator, DISCO-TAB couples a fine-tuned Large Language Model (LLM) with a hierarchical reinforcement learning (RL) optimization strategy, enabling the generator to receive dense, structured feedback aligned with clinical reasoning. Specifically, we design a multi-objective discriminator ensemble that evaluates generated records at four complementary granularities: \emph{token-level} constraints enforce syntactic correctness of numerical values and categorical codes; \emph{sentence-level} constraints penalize local semantic contradictions (e.g., demographic--diagnosis inconsistencies); \emph{feature-level} constraints preserve statistically significant clinical dependencies automatically discovered from real EHR data; and \emph{row-level} constraints enforce global coherence of entire patient profiles. 

To further address the endemic class imbalance of biomedical datasets, we introduce \emph{Inverse-Frequency Reward Shaping} (IFRS), which dynamically scales reinforcement signals to incentivize faithful generation of rare but clinically critical patient subpopulations. This hierarchical, reward-shaped RL formulation allows DISCO-TAB to localize and correct violations at the appropriate semantic level during generation, rather than filtering errors post hoc. As a result, the framework simultaneously improves minority-class coverage, preserves fine-grained feature correlations, and suppresses medically implausible hallucinations. Extensive evaluation across ten diverse benchmarks demonstrates that this design translates directly into practice: DISCO-TAB achieves state-of-the-art downstream utility, improving train-on-synthetic, test-on-real performance by up to 12.3\% over GAN, diffusion, and agentic LLM baselines, while maintaining high statistical fidelity (JSD $< 0.01$) and robust privacy guarantees.
These results demonstrate that explicitly encoding clinical constraints into the optimization loop is essential for trustworthy biomedical data synthesis, particularly in small-sample and high-imbalance regimes where existing generative models fail.

\noindent \textbf{Our Contributions.} 
are as follows:

1. \textit{Hierarchical Clinical Verification:} We propose a novel multi-granular discriminator system that enforces validity from the atomic token level (e.g., valid lab values) to the holistic record level, significantly reducing medical hallucinations.

2. \textit{Automated Constraint Discovery:} We introduce an automated pre-processing module that extracts latent statistical dependencies from raw EHR data, initializing feature-specific discriminators without requiring manual rule engineering by clinicians.

3. \textit{SOTA Performance on Medical Benchmarks:} We rigorously validate DISCO-TAB across nine datasets, including high-dimensional, small-sample medical benchmarks. Our framework achieves up to 12.3\% improvement in downstream classifier utility over GAN and Diffusion baselines while maintaining exceptional statistical fidelity.

%% file: 03-relatedwork.tex
Synthetic tabular data generation has progressed from classical parametric modeling and oversampling to deep generative frameworks that aim to approximate high-dimensional mixed-type joint distributions under privacy and utility constraints. Following the positioning of DISCO-TAB in the current manuscript, we organize prior work along five complementary axes: (i) adversarial and auto-encoding methods, (ii) diffusion-based synthesis, (iii) autoregressive Transformer/LLM-based synthesis, (iv) constraint- and feedback-driven optimization, and (v) privacy evaluation and attack models.

This categorization reflects two orthogonal design dimensions that have emerged in the tabular synthesis literature. The first dimension concerns the \emph{generative backbone}, i.e., the probabilistic mechanism used to model the joint distribution of heterogeneous features. The second dimension captures the \emph{mechanisms used to enforce validity, control, and trustworthiness}, encompassing explicit structural constraints and feedback-driven optimization. We adopt this dual-axis organization to highlight that high-fidelity biomedical synthesis requires not only a powerful generative model, but also principled control mechanisms that govern semantic validity, minority-class coverage, and privacy risk, an architectural separation that directly motivates the design of DISCO-TAB.

\subsection{Adversarial and Auto-Encoder Methods}
Early deep tabular synthesizers adapted GAN training to heterogeneous columns via conditional vectors. CTGAN~\cite{xu2019modeling} remains a canonical baseline. In healthcare, MedGAN~\cite{choi2017generating} combined autoencoders with adversarial learning to handle high-dimensional discrete codes.
More recently, TAEGAN \cite{li2025taegangeneratingsynthetictabular} revisited the GAN architecture by incorporating a masked auto-encoder as the generator. While TAEGAN achieves high utility through efficient adversarial training, it remains fundamentally a GAN-based approach, which often struggles to capture the long-range semantic dependencies (e.g., patient trajectories) that language models naturally encode.

\subsection{Diffusion Models for Tabular Synthesis}
Diffusion probabilistic models, such as TabDDPM~\cite{kotelnikov2023tabddpm} and TabSyn \cite{tabsyn}, have set new standards for privacy by modeling data via iterative denoising. However, a recurring limitation for clinical data is that the Gaussian noise assumption can blur fine-grained thresholds (e.g., specific blood pressure cutoffs) and rare co-morbidity structures. As shown in our privacy analysis (Section \ref{sec:privacyResults}), this often leads to privacy through noise, achieving high metrics by drifting away from the valid clinical manifold.

\subsection{Autoregressive Transformers and LLM-Based Baselines}
Treating tabular synthesis as a sequence modeling problem has gained traction. Fine-tuning approaches such as GReaT~\cite{borisov2022language}, RealTabFormer~\cite{solatorio2023realtabformer}, and HARMONIC \cite{Wang2024HARMONICHL} demonstrate that Transformer-based generators can capture global distributions by linearizing records into token sequences. Parallel to fine-tuning, prompt-based strategies like SynLLM \cite{ilaty2025synllm} and AIGT \cite{zhang2024aigtaigenerativetable} utilize context engineering to guide pre-trained LLMs in structured data generation without extensive weight updates. 
Focusing on architectural efficiency, TabularARGN \cite{sidorenko2025tabularargn} recently proposed a randomized permutation framework for fast conditional generation.
\textit{Limitation:} While these autoregressive approaches excel at semantic modeling, they rely largely on likelihood maximization. Consequently, they risk generating samples that are statistically plausible but semantically inconsistent (e.g., demographic--diagnosis conflicts), as they lack explicit mechanisms to enforce complex medical domain logic during the generation process.

\subsection{Constraint- and Feedback-Driven Optimization}
Beyond choosing a backbone, enforcing validity requires explicit control mechanisms. In the database community, approaches such as Kamino~\cite{ge2021kamino} and PrivMRF~\cite{cai2021privmrf} synthesize data while preserving integrity constraints. However, these are often rigid and difficult to scale to high-dimensional generative models.

In the neural domain, recent works have attempted to guide generation via feedback. TAGAL \cite{ronval2025tagal} employs an agentic loop where an LLM critiques its own outputs via textual prompts. However, this relies on the model's inherent reasoning and lacks mechanisms to enforce hard structural constraints. Similarly, P-TA \cite{yang2025ptausingproximalpolicy} applies Proximal Policy Optimization (PPO) with a binary discriminator. While this introduces reinforcement learning, it relies on a sparse scalar reward signal that cannot distinguish between syntactic errors and semantic inconsistencies. DISCO-TAB addresses these limitations by decomposing the reward signal into hierarchical, granular components that explicitly optimize for distinct levels of clinical validity.

\subsection{Privacy Evaluation and Attacks on Synthetic Tabular Releases}
A critical practical dimension is that synthetic data is not automatically private; generative models can leak training information via memorization or distributional collisions.
TableGAN-MCA~\cite{hu2021tableganmca} introduces a \emph{membership collision} threat model showing that, even when classic membership inference appears weak, black-box access to synthetic samples can enable partial recovery of training records due to collisions with real data points.
This line of work motivates stronger privacy evaluation for tabular synthesis beyond utility-only metrics. Differential privacy (DP) mechanisms for synthetic data generation have also been widely studied; for example, PATE-GAN~\cite{jordon2019pategan} provides DP guarantees for GAN-based synthesis through private aggregation of teacher discriminators.
In domains with high sensitivity (e.g., EHRs), these results collectively emphasize that trustworthy synthesis requires \emph{both} (i) structured validity/utility preservation and (ii) explicit privacy auditing against modern attacks.

\noindent\textbf{Summary and Positioning.}
Prior GAN and diffusion paradigms provide strong baselines for mixed-type fidelity, and Transformer/LLM approaches improve semantic modeling via tokenization and pretraining. However, the literature also indicates two persistent gaps: (1) enforcing \emph{hierarchical clinical/structural validity} beyond scalar objectives, and (2) sustaining \emph{minority-class coverage} without mode dropping, while simultaneously remaining robust under modern privacy attack models. DISCO-TAB is designed to address these gaps by combining multi-objective, multi-granular discriminators with reward-shaped RL fine-tuning and by evaluating utility--fidelity--privacy jointly.

%% file: 04-methodology.tex
We propose \textbf{DISCO-TAB}, a hierarchical reinforcement learning framework for synthesizing high-fidelity, \emph{privacy-aware} tabular data. The proposed approach mitigates the stochastic and unconstrained nature of standard large language models (LLMs) by explicitly enforcing structural, semantic, and statistical validity through a multi-granular feedback mechanism. Rather than relying on implicit likelihood optimization, DISCO-TAB formulates data generation as a constrained sequential decision-making problem guided by interpretable validity signals.

\begin{figure*}[t]
    \centering
    \includegraphics[width=\columnwidth]{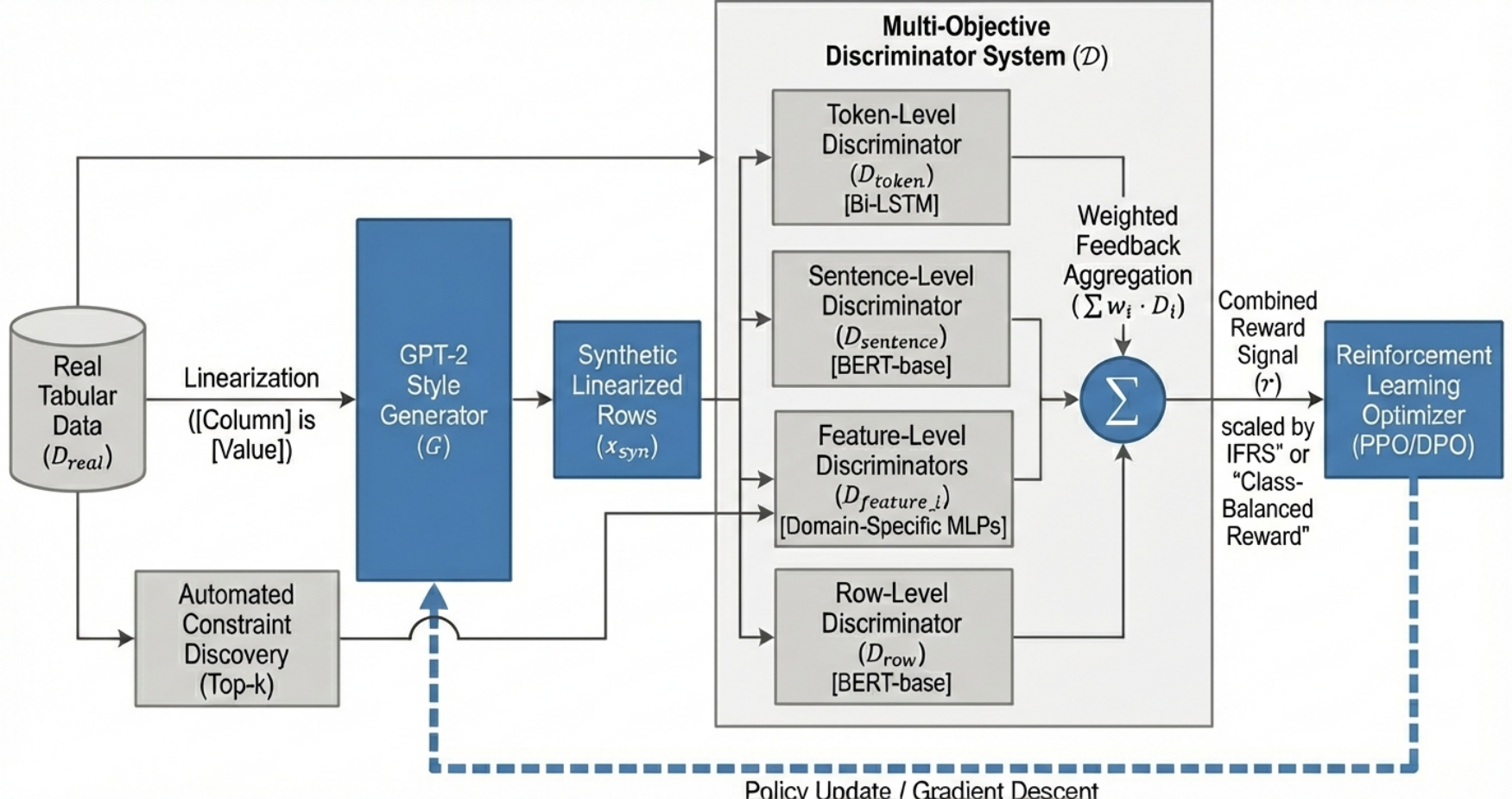} 
    \caption{\textbf{The DISCO-TAB framework for tabular data synthesis.}
\textbf{(A) Automated Constraint Discovery:} The system analyzes the real dataset ($\mathcal{D}_{real}$) to compute a global correlation matrix and identify statistically significant feature dependencies (e.g., Age vs. Comorbidities), which are used to initialize the feature-level discriminator ($D_{feat}$).
\textbf{(B) Hierarchical Discriminators:} A multi-objective discriminator ensemble evaluates generated records ($S$) at four semantic granularities: token-level (syntactic validity), sentence-level (local semantic plausibility), feature-level (dependency consistency), and row-level (global coherence).
\textbf{(C) Inverse Frequency Reward Shaping (IFRS):} During reinforcement learning, the PPO reward is scaled based on the rarity of the target class associated with the generated record, mitigating mode collapse in imbalanced datasets.}
    \label{fig:framework}
\end{figure*}

As illustrated in Fig.~\ref{fig:framework}, the framework consists of three tightly integrated components: (1) an \emph{Automated Constraint Discovery} module that identifies latent clinical and statistical dependencies directly from real data; (2) a \emph{Hierarchical Discriminator System} that evaluates generated samples at four semantic resolutions, token, sentence, feature, and row, to localize and penalize distinct forms of invalidity; and (3) a \emph{Reward-Shaped Proximal Policy Optimization (PPO)} procedure that aligns the generative policy with both global distributional fidelity and robust minority-class representation.

DISCO-TAB does not enforce formal differential privacy guarantees during training; instead, privacy is assessed empirically using standard auditing metrics, including distance-based similarity measures and membership inference resistance, as detailed in Section~\ref{sec:privacyResults}.
Collectively, these components enable DISCO-TAB to generate synthetic tabular data that is statistically faithful, structurally coherent, and suitable for downstream clinical analytics while maintaining strong empirical privacy properties.

Throughout this section, the generator $G_\theta$ and policy $\pi_\theta$ refer to the same autoregressive model, viewed respectively as a data generator and as a stochastic policy; $\mathcal{D}_{ens}$ denotes the discriminator ensemble, with $D_k(\cdot)$ indicating an individual discriminator operating at granularity $k$.

\subsection{Problem Formulation and Linearization}\label{subSec:PrbFrm}

We use $S_i$ to denote a specific serialized record corresponding to row $r_i \in \mathcal{D}_{real}$, and $S$ to denote a generic generated sequence sampled from the policy.
Let $\mathcal{D}_{real} = \{r_1, r_2, \dots, r_N\}$ be a private clinical dataset consisting of $N$ patient records. Each record $r_i$ contains $M$ heterogeneous features $\mathcal{F} = \{f_1, f_2, \dots, f_M\}$, where $f_j$ may be continuous (e.g., BMI), categorical (e.g., Diagnosis Code), or ordinal (e.g., Disease Stage).

To leverage the semantic reasoning of pre-trained LLMs, we employ a schema-aware linearization function $\mathcal{L}: \mathcal{R}^M \rightarrow \mathcal{V}^*$ that transforms a structured row $r_i$ into a natural language sequence $S_i = \{w_1, w_2, \dots, w_T\}$. We adopt a verbose template strategy to preserve semantic context:
\begin{equation}
    S_i = \text{``[}f_1\text{] is [}v_{i,1}\text{], [}f_2\text{] is [}v_{i,2}\text{],} \dots \text{"}
\end{equation}
The generator $G_\theta$, initialized with a GPT-2 backbone, models the probability of a sequence as $P_\theta(S) = \prod_{t=1}^T P(w_t | w_{<t})$.

To ensure reproducibility and prevent schema drift, we employ a deterministic linearization scheme shared across all datasets. Feature names are treated as fixed lexical tokens and appear in a predefined order consistent with the original tabular schema. Continuous values are standardized using statistics computed on $\mathcal{D}_{real}$ and rendered with fixed decimal precision. Categorical and ordinal values are represented using canonical string identifiers derived from the original data vocabulary. Missing values are explicitly encoded using a reserved \texttt{[MISSING]} token to distinguish absence from valid zero or null-equivalent values. This schema-aware representation enforces consistent structure during generation and enables discriminators to localize syntactic and semantic violations reliably.

\subsection{Automated Constraint Discovery}
Medical datasets often contain strong statistical dependencies across heterogeneous variables (e.g., Age vs. Comorbidities) that are critical for clinical validity.
Relying on manual rule definitions is unscalable. Instead, we introduce an unsupervised constraint discovery phase prior to training.

We compute a global correlation matrix $\mathbf{C} \in \mathbb{R}^{M \times M}$ for $\mathcal{D}_{real}$. We utilize Pearson correlation for numerical-numerical pairs, Cramér's V for categorical-categorical pairs, and the Correlation Ratio for mixed types.
We identify the set of critical dependency pairs $\mathcal{P}_{crit}$ by first filtering feature pairs whose association strength exceeds a predefined threshold $\delta_{thresh}$, and then selecting the top-$k$ strongest pairs among them:

\begin{equation}
    \mathcal{P}_{crit} = \text{Top-}k \Big( \{ (f_a, f_b) \mid |\mathbf{C}_{ab}| \geq \delta_{thresh} \} \Big)
\end{equation}
These discovered pairs initialize the input heads of our Feature-Level Discriminator ($D_{feat}$), ensuring the model explicitly optimizes for the preservation of these specific, clinically significant relationships.
This two-stage selection ensures that only statistically meaningful dependencies are retained while bounding the number of feature-level constraints for computational efficiency and training stability.

While this discovery process is correlation-based, it is sufficient to surface dominant linear and mixed-type associations that capture clinically meaningful structure without requiring explicit causal or high-order dependency modeling.

\subsection{Hierarchical Multi-Objective Discriminators}
To guide the generator towards clinically valid distributions, we employ a composite discriminator system $\mathcal{D}_{ens}$ operating at four granularities.

Table~\ref{tab:discriminator_overview} summarizes the hierarchical discriminator ensemble, highlighting the semantic granularity, inputs, and roles of each component.


\subsubsection{Level 1: Token Consistency ($D_{token}$)}
Medical data requires strict syntactic adherence (e.g., ICD codes must follow alphanumeric formats; floats must not contain characters). We employ a lightweight Bidirectional LSTM to evaluate the syntactic validity of each generated token $w_t$ given its local context. This provides immediate, low-level feedback to prevent syntax errors.

We adopt a lightweight Bidirectional LSTM for $D_{token}$ rather than deterministic rule-based validation (e.g., regular expressions or hard-coded type checks) for two reasons. First, learned sequence models generalize across heterogeneous schemas and tokenization patterns without requiring manual specification of dataset-specific rules. Second, contextual modeling enables $D_{token}$ to detect subtle syntactic violations that depend on local token context (e.g., malformed numeric values or invalid code prefixes) while remaining differentiable and seamlessly integrable into the reinforcement learning objective. This design balances expressiveness and computational efficiency, providing robust low-level validity signals without constraining the generator through brittle decoding rules.

\subsubsection{Level 2: Semantic Plausibility ($D_{sent}$)}

In the context of serialized tabular data, a ``sentence'' corresponds to a single feature--value clause generated by the linearization template (e.g., ``\texttt{Gender is Male}'' or ``\texttt{Pregnancy is Yes}''). Each clause is delimited deterministically based on the schema-aware serialization rules described in Section~\ref{subSec:PrbFrm}. During training and inference, $D_{sent}$ operates on individual clauses rather than full records, enabling localized detection of semantic inconsistencies between a feature and its assigned value without confounding effects from unrelated attributes.

Local contradictions (e.g., "Gender: Male, Pregnancy: Yes") undermine clinical trust. We utilize a BERT-base encoder to evaluate short variable-value segments. This discriminator leverages pre-trained semantic knowledge to penalize logical hallucinations that are syntactically correct but semantically impossible.

\subsubsection{Level 3: Feature Correlation ($D_{feat}$)}
For every pair $(f_a, f_b) \in \mathcal{P}_{crit}$, we extract the contextual embeddings $h_a, h_b$ from the generator's last hidden layer. 

For each feature--value clause in the serialized sequence, we maintain a deterministic mapping between feature identifiers and token spans based on the schema-aware linearization rules. Let $\mathcal{T}(f_a)$ denote the set of token indices corresponding to the \emph{value} of feature $f_a$. We compute the feature embedding $h_a$ by mean-pooling the generator's final-layer hidden states over these value tokens:
\begin{equation}
h_a = \frac{1}{|\mathcal{T}(f_a)|} \sum_{t \in \mathcal{T}(f_a)} H_t,
\end{equation}
where $H_t$ denotes the contextual embedding at token position $t$. Feature-name tokens are excluded from pooling to prevent the discriminator from exploiting trivial lexical cues. The same procedure is applied to obtain $h_b$ for feature $f_b$.
These embeddings are concatenated and passed through a specialized \textbf{Multi-Layer Perceptron (MLP)} projection head:
\begin{equation}
    Score_{feat} = \sigma(W_2 \cdot \text{ReLU}(W_1 [h_a \oplus h_b] + b_1) + b_2)
\end{equation}
This encourages preservation of these dependencies if the joint distribution of highly correlated clinical features deviates from the real data manifold.
The output of this projection head is treated as the feature-level discriminator score, and $D_{feat}(S)$ is obtained by aggregating these scores across all $(f_a,f_b)\in\mathcal{P}_{crit}$ for the record.

\subsubsection{Level 4: Global Coherence ($D_{row}$)}
To assess holistic clinical plausibility, the row-level discriminator evaluates entire serialized records. Specifically, $D_{row}$ operates on the sequence of final-layer token embeddings produced by the generator. These embeddings are aggregated using mean pooling across all token positions to obtain a fixed-length representation of the record, which is then passed through a lightweight Multi-Layer Perceptron (MLP) to produce a scalar realism score:
\begin{equation}
D_{row}(S_i) = \sigma\!\left(\text{MLP}\!\left(\frac{1}{T}\sum_{t=1}^{T} H_t\right)\right),
\end{equation}
where $H_t$ denotes the contextual embedding at position $t$ and $T$ is the sequence length. This design captures global feature interactions while remaining computationally efficient and independent of other discriminators, enabling scalable training without introducing heavy Transformer-based overhead.

\begin{algorithm}[t]
\footnotesize
\captionsetup{font=footnotesize}
\caption{DISCO-TAB Training Procedure for Clinical Data}
\label{alg:discotab}
\begin{algorithmic}[1] 
\Require Real Data $\mathcal{D}_{real}$, Pre-trained Generator $G_{\theta_{init}}$, Correlation Threshold $\delta_{thresh}$, Max Dependency Count $k$, Balancing Factor $\alpha$
\Ensure Optimized Policy $\pi_{\theta^*}$

\item[] \textbf{Phase 1: Constraint Discovery} 
\State $\mathbf{C} \gets \text{ComputeGlobalCorrelation}(\mathcal{D}_{real})$
\State $\mathcal{P}_{crit} \gets \text{ExtractPairs}(\mathbf{C}, \delta_{thresh}, k)$ \Comment{Identify dependencies}
\State Initialize Discriminator Ensemble $\mathcal{D}_{ens} = \{D_{token}, D_{sent}, D_{feat}, D_{row}\}$

\item[] \textbf{Phase 2: RL Optimization Loop}
\For{epoch $e = 1$ to $E$}
    \State $\mathcal{B}_{syn} \gets \text{GenerateBatch}(G_{\theta})$
    \State $\mathcal{B}_{real} \gets \text{SampleBatch}(\mathcal{D}_{real})$
    
    \State \textit{// Step A: Update Discriminators}
    \State $\mathcal{L}_{disc} \gets \sum_k \mu_k \mathcal{L}_{disc}^{(k)}$
    \State Update $\mathcal{D}_{ens}$ to minimize $\mathcal{L}_{disc}$
    
    \State \textit{// Step B: Policy Update with IFRS}
    \For{sequence $S \in \mathcal{B}_{syn}$}
        \State $y \gets \text{ExtractDiagnosisLabel}(S)$
        \State $W_{y} \gets 1 + \alpha \cdot \max(0, \log(\frac{N_{total}}{N_y \cdot |\mathcal{Y}|}))$ \Comment{Calculate Rarity Weight}
        \State $R_{raw} \gets \sum \lambda_k D_k(S)$
        \State $R_{total} \gets R_{raw} \cdot W_{y}$ \Comment{Apply Inverse Frequency Shaping}
    \EndFor
    
    \State Update policy and value parameters $(\theta,\phi)$ by maximizing $\mathcal{L}(\theta,\phi)$ via PPO
\EndFor
\State \textbf{return} $\pi_{\theta^*}$
\end{algorithmic}
\end{algorithm}

\subsubsection{Discriminator Training and Loss Aggregation}

We use $\mathcal{B}_{real}^{(k)}$ and $\mathcal{B}_{syn}^{(k)}$ to denote the real and synthetic minibatches constructed at discriminator granularity $k$.
Each discriminator in the ensemble $\mathcal{D}_{ens} = \{D_{token}, D_{sent}, D_{feat}, D_{row}\}$ is trained as a binary classifier to distinguish real samples from synthetic ones at its respective granularity. Let $\mathcal{B}_{real}$ and $\mathcal{B}_{syn}$ denote minibatches of real and generated records, respectively.

For both real and synthetic samples, token embeddings $H_t$ are obtained via a forward pass through the generator $G_\theta$; for real sequences, embeddings are computed under teacher forcing using the ground-truth token order.

\noindent\textbf{Granularity-Specific Training Sets.}
For each discriminator $D_k$, we construct a task-specific dataset:

1. $D_{token}$ operates on individual tokens extracted from serialized records.

2. $D_{sent}$ operates on short variable--value segments parsed from the sequence.

3.  $D_{feat}$ operates on embedding pairs $(h_a,h_b)$ corresponding to feature pairs $(f_a,f_b)\in\mathcal{P}_{crit}$.

4.  $D_{row}$ operates on full serialized records $S$.

Each sample is labeled as real ($y=1$) if drawn from $\mathcal{D}_{real}$ and synthetic ($y=0$) otherwise.

For discriminator $D_k$, we define a binary cross-entropy loss:
\begin{equation}
\mathcal{L}_{disc}^{(k)} =
\mathbb{E}_{x \sim \mathcal{B}_{real}^{(k)}}\!\left[-\log D_k(x)\right]
+
\mathbb{E}_{x \sim \mathcal{B}_{syn}^{(k)}}\!\left[-\log(1 - D_k(x))\right],
\end{equation}
where $\mathcal{B}_{real}^{(k)}$ and $\mathcal{B}_{syn}^{(k)}$ denote the real and synthetic samples at granularity $k$.

The overall discriminator loss is the weighted sum of the individual losses:
\begin{equation}
\mathcal{L}_{disc} = \sum_{k \in \{token,\,sent,\,feat,\,row\}} \mu_k \, \mathcal{L}_{disc}^{(k)},
\end{equation}
where $\mu_k$ controls the relative contribution of each discriminator during training. In our experiments, we set $\mu_k$ uniformly unless stated otherwise.

\subsection{Optimization via Reward-Shaped RL}
\label{sec:rl_opt}

We cast tabular record synthesis as an episodic Markov Decision Process (MDP) in which the generator is an autoregressive policy. For a given record template, the policy $\pi_\theta$ generates a serialized sequence $S = (w_1,\dots,w_T)$ token-by-token.
At decoding step $t$, the \emph{state} is the prefix $s_t = (w_1,\dots,w_{t-1})$, the \emph{action} is the next-token emission $a_t = w_t$, and the episode terminates at $t=T$ when an end-of-record token is produced. Thus, the trajectory distribution is $S \sim \pi_\theta(\cdot)$ with $\pi_\theta(S)=\prod_{t=1}^{T}\pi_\theta(w_t \mid w_{<t})$.

\noindent\textbf{Reward Construction and Timing.}
After generating a complete sequence $S$, DISCO-TAB computes a composite reward from the discriminator ensemble. While the discriminators operate at different granularities (token/sentence/feature/row), we use their outputs to form a single terminal reward assigned to the trajectory:
\begin{equation}
R_{\text{raw}}(S) = \sum_{k \in \{token,\,sent,\,feat,\,row\}} \lambda_k\, D_k(S),
\end{equation}
where $D_k(\cdot)\in[0,1]$ denotes the score produced by discriminator $k$ (aggregated over tokens/segments/pairs as appropriate) and $\lambda_k$ controls its contribution. This design yields a dense \emph{evaluation signal} (via multiple discriminators) while preserving a simple episodic RL objective. In practice, the resulting terminal reward is broadcast to all time steps through the return used for policy-gradient updates.

\noindent\textbf{Discriminator Output Aggregation.}
Each discriminator produces multiple local scores at its operating granularity, which are aggregated into a scalar record-level score $D_k(S)$. Specifically, $D_{token}(S)$ and $D_{sent}(S)$ are computed as the mean score over all tokens and feature--value clauses in $S$, respectively; $D_{feat}(S)$ is computed as the mean score over all feature pairs $(f_a,f_b)\in\mathcal{P}_{crit}$ evaluated for the record; and $D_{row}(S)$ directly outputs a scalar score for the entire sequence. Mean aggregation was chosen for stability and to avoid dominance by any single local violation.

\subsubsection{Inverse Frequency Reward Shaping (IFRS)}
Clinical datasets are often severely imbalanced; maximum-likelihood training can underrepresent minority cohorts. To counter this, we scale the reward by a class-dependent weight. For a generated record with target label $y \in \mathcal{Y}$, we define:
\begin{equation}
W_y = 1 + \alpha \cdot \max\left(0, \log\left(\frac{N_{\text{total}}}{N_y \cdot |\mathcal{Y}|}\right)\right),
\end{equation}
where $N_y$ is the frequency of class $y$ in $\mathcal{D}_{real}$, $N_{\text{total}}=\sum_{y\in\mathcal{Y}}N_y$, and $\alpha$ is a scaling hyperparameter. The final shaped reward is:
\begin{equation}
R_{\text{total}}(S) = W_y \cdot R_{\text{raw}}(S).
\end{equation}

\subsubsection{PPO Objective with KL Regularization}
We optimize the generator parameters $\theta$ using Proximal Policy Optimization (PPO). Let $\pi_{\theta_{\text{old}}}$ denote the policy before the update, and define the likelihood ratio
\begin{equation}
r_t(\theta) = \frac{\pi_\theta(w_t \mid w_{<t})}{\pi_{\theta_{\text{old}}}(w_t \mid w_{<t})}.
\end{equation}
We employ a learned value function $V_\phi(s_t)$ to reduce variance and compute advantages. Using the terminal reward, we define the Monte-Carlo return $\hat{G}_t = R_{\text{total}}(S)$ for all $t$ and the advantage
\begin{equation}
\hat{A}_t = \hat{G}_t - V_\phi(s_t).
\end{equation}
The PPO clipped surrogate objective is:
\begin{equation}
\mathcal{L}_{\text{PPO}}(\theta) = \mathbb{E}\left[
\min\left(r_t(\theta)\hat{A}_t,\;
\text{clip}(r_t(\theta), 1-\epsilon, 1+\epsilon)\hat{A}_t\right)
\right],
\end{equation}
where $\epsilon$ is the clipping threshold. To prevent the policy from drifting excessively from the reference model, we additionally include KL regularization to a fixed reference policy $\pi_{\text{ref}}$ (initialized from the pre-trained backbone):
\begin{equation}
\begin{aligned}
\mathcal{L}(\theta,\phi) =
&\;\mathcal{L}_{\text{PPO}}(\theta)
- \beta\,\mathbb{E}\!\left[
\text{KL}\!\left(\pi_\theta(\cdot \mid s_t)\,\|\,\pi_{\text{ref}}(\cdot \mid s_t)\right)
\right] \\
&\; - c_v\,\mathbb{E}\!\left[
\left(V_\phi(s_t)-\hat{G}_t\right)^2
\right].
\end{aligned}
\end{equation}

where $\beta$ controls the KL penalty and $c_v$ weights the value-function regression loss. This formulation ensures stable optimization while aligning the generator with multi-granular validity constraints and minority-class coverage.

\subsubsection{Mitigation of Reward Exploitation}
Reinforcement learning–based generative models are susceptible to reward exploitation, where the policy discovers degenerate strategies that satisfy a subset of objectives without producing semantically meaningful outputs. DISCO-TAB mitigates this risk through several complementary design choices. First, the KL regularization term constrains policy updates to remain close to the reference pre-trained model, preventing abrupt divergence toward adversarial or nonsensical generations. Second, reward signals are derived from a multi-objective discriminator ensemble operating at distinct semantic granularities, making it difficult for the generator to exploit a single weak signal in isolation. Third, rewards are assigned at the trajectory level rather than per-token, discouraging locally optimal but globally invalid sequences. Finally, discriminators are updated continuously using both real and synthetic samples, reducing the risk of stale or easily exploitable decision boundaries. Together, these mechanisms promote stable training and ensure that improvements in reward correspond to meaningful gains in clinical validity and data utility.

\subsection{Training Configuration and Hyperparameters}
To ensure reproducibility, we briefly summarize the key training hyperparameters used in DISCO-TAB. Unless stated otherwise, hyperparameters are held fixed across datasets.

\noindent\textbf{Generator and PPO Optimization.}
The generator is initialized from a pre-trained GPT-2 backbone. PPO is trained with a fixed learning rate, clipping parameter $\epsilon$, and KL-regularization weight $\beta$ to control deviation from the reference policy. A learned value function is jointly optimized using mean-squared error loss weighted by coefficient $c_v$. All PPO-related hyperparameters, including learning rates, batch sizes, and number of update steps per epoch, are reported in Section~\ref{sec:experiments}.

\noindent\textbf{Discriminator Training.}
Each discriminator in $\mathcal{D}_{ens}$ is trained using binary cross-entropy loss with equal weighting ($\mu_k = 1$) unless otherwise specified. Discriminators are updated for a fixed number of steps per epoch using minibatches of real and synthetic samples drawn at their respective granularities. The relative contribution of each discriminator to the generator reward is controlled by $\lambda_k$, which is set uniformly in all experiments.

\noindent\textbf{Constraint Discovery and Reward Shaping.}
The constraint discovery threshold $\delta_{thresh}$ and the maximum number of retained dependency pairs $k$ are fixed across datasets. The Inverse Frequency Reward Shaping parameter $\alpha$ is selected from a small validation sweep and then held constant. Full hyperparameter values are provided in the experimental setup for completeness.

%% file: 05-result.tex
To rigorously evaluate the efficacy of DISCO-TAB, we conducted a comprehensive benchmarking study against seven state-of-the-art generative frameworks across a diverse suite of nine datasets spanning medical, financial, and social domains. Our analysis goes beyond simple leaderboard reporting; we aim to dissect the mechanistic reasons behind performance differentials, specifically examining how hierarchical reinforcement learning resolves the traditional trade-off between statistical fidelity and sample diversity.

\subsection{Experimental Setup}

We evaluate DISCO-TAB on a diverse suite of benchmarks designed to span multiple data regimes, ranging from small, high-dimensional medical datasets such as Heart Failure ($N=299, D=13$) to large-scale, severely imbalanced datasets such as Diabetes ($N=100{,}000$, imbalance ratio $1{:}9$). This selection intentionally stresses complementary failure modes of existing generative models: autoregressive Transformers typically underperform in data-scarce settings, while adversarial models are prone to mode collapse under extreme class imbalance.

DISCO-TAB is compared against representative state-of-the-art baselines from all major paradigms of tabular data synthesis, including adversarial approaches (CTGAN~\cite{xu2019modeling}, TAEGAN~\cite{li2025taegangeneratingsynthetictabular}), autoregressive Transformer-based models (RealTabFormer~\cite{solatorio2023realtabformer} and the commercial AutoML system MostlyAI ~\cite{mostlyai}), diffusion and latent-variable methods (TabDDPM~\cite{kotelnikov2023tabddpm}, TabSyn~\cite{tabsyn}), and recent agentic LLM-based generation frameworks (TAGAL~\cite{ronval2025tagal}).

\subsection{Machine Learning Utility Analysis}

The primary measure of synthetic data value is its Train on Synthetic, Test on Real (TSTR) performance. Table \ref{tab:main_results} presents the F1-scores of Random Forest classifiers trained on generated data with 5-fold cross-validation. Hyperparameters were standardized across all experiments to ensure fair comparison between real and synthetic datasets.

\noindent\textbf{Relative Utility Gain ($\Delta$) Definition and Interpretation.}
To explicitly quantify the performance margin of DISCO-TAB over competing generative models, we define the relative utility gain for each dataset $d$ as
\begin{equation}
\Delta_d \;=\; F1_{\text{DISCO-TAB}}^{(d)} \;-\; \max_{m \in \mathcal{B}} F1_m^{(d)},
\end{equation}
where $\mathcal{B}$ denotes the set of all baseline methods excluding DISCO-TAB. 
Positive values of $\Delta_d$ indicate that DISCO-TAB achieves strictly higher downstream utility than all competing baselines on dataset $d$, whereas negative values indicate that at least one baseline attains superior performance.

\noindent\textbf{Medical vs.\ Non-Medical Performance Regimes.}
This $\Delta$-based analysis reveals a clear and domain-consistent performance pattern. 
Across the six medical benchmarks (Heart Failure, Breast Cancer, Liver Disorders, Parkinsons, Obesity, and Diabetes), DISCO-TAB achieves positive performance gains on \emph{five out of six} datasets, with particularly pronounced improvements on small-sample and highly imbalanced clinical datasets. 
The largest margins are observed on Liver Disorders ($\Delta=+0.382$) and Heart Failure ($\Delta=+0.123$), directly supporting our central hypothesis that hierarchical reinforcement learning with multi-granular discriminators is especially effective in data-scarce medical regimes. 
The only exception is the Diabetes dataset, where a large-scale commercial AutoML baseline marginally outperforms DISCO-TAB, reflecting the reduced importance of structural and minority-class constraints in extremely large datasets.

In contrast, on the non-medical benchmarks (German Credit, Bank Marketing, and Census Income), which are included to evaluate cross-domain generalizability, DISCO-TAB remains competitive but does not consistently dominate specialized industrial baselines. While a substantial gain is achieved on German Credit ($\Delta=+0.275$), modest negative margins are observed on Bank Marketing ($\Delta=-0.014$) and Census Income ($\Delta=-0.006$). This trend indicates that DISCO-TAB’s primary advantages emerge in domains where semantic validity, structural coherence, and minority-class preservation are critical, rather than in large, well-balanced tabular datasets.

Traditional deep learning methods typically require large training corpora to converge reliably. On the Heart Failure dataset ($N=299$), baseline methods such as RealTabFormer and TabDDPM fail to learn a meaningful decision boundary, achieving F1-scores close to random guessing ($\approx 0.55$--$0.60$). Even the commercial AutoML solution MostlyAI exhibits limited utility (0.598). In stark contrast, DISCO-TAB attains near-perfect downstream performance ($F1=1.000$, $\Delta=+0.123$).
This result substantiates our hypothesis regarding \emph{Hierarchical Feedback Efficiency}. While standard autoregressive training objectives based on cross-entropy provide sparse and indirect learning signals in small-data regimes, DISCO-TAB leverages dense, multi-granular reinforcement signals at the token, feature, and row levels. As a result, the generator is explicitly guided to remain on the valid data manifold, mitigating the mode-averaged noise commonly observed in unguided Transformer-based synthesis.

\noindent\textbf{Overcoming Mode Collapse in Complex Manifolds.}
The Obesity dataset poses a challenging multi-class classification problem characterized by severe class imbalance. Adversarial baselines collapse in this setting (CTGAN: 0.105; TAEGAN: 0.215), consistent with the vanishing-gradient pathology of GAN training on discrete and imbalanced data. DISCO-TAB achieves an F1-score of 0.929 ($\Delta=+0.027$), substantially outperforming both GAN-based and agentic LLM baselines such as TAGAL (0.790).
This improvement is driven by our Inverse Frequency Reward Shaping (IFRS) mechanism, which dynamically amplifies reinforcement signals for minority-class generation. By integrating class-aware rewards directly into the RL objective, DISCO-TAB counteracts the tendency of large language models to overfit dominant classes, enabling robust synthesis of rare but clinically significant patterns.

\begin{table*}[t]
\centering
\caption{Downstream ML utility (F1-score) under the Train-on-Synthetic, Test-on-Real (TSTR) protocol. 
$\Delta$ denotes the absolute percentage point differences  between DISCO-TAB and the strongest baseline per dataset 
(positive indicates DISCO-TAB superiority; negative indicates a stronger baseline). 
The first six datasets are medical benchmarks, while the last three (German Credit, Bank Marketing, Census Income) are non-medical datasets included to evaluate cross-domain generalizability.}
\label{tab:main_results}
\begin{tabular}{l|ccccccc|cc}
\toprule
\textbf{Dataset} & \textbf{CTGAN} & \textbf{TAEGAN} & \textbf{TabDDPM} & \textbf{TabSyn} & \textbf{RealTabFormer} & \textbf{MostlyAI} & \textbf{TAGAL} & \textbf{DISCO-TAB} & $\Delta$ \\ 
\midrule
Heart Failure   & 0.778 & 0.655 & 0.665 & 0.877 & 0.548 & 0.598 & 0.839 & \textbf{1.000} & +0.123 \\
Breast Cancer   & 0.549 & 0.693 & 0.689 & 0.934 & 0.495 & 0.498 & 0.990 & \textbf{0.994} & +0.004 \\
Liver Disorders & 0.496 & 0.576 & 0.133 & 0.510 & 0.528 & 0.574 & 0.608 & \textbf{0.990} & +0.382 \\
Parkinsons      & 0.594 & 0.595 & 0.637 & 0.610 & 0.637 & 0.637 & 0.886 & \textbf{0.966} & +0.080 \\
Obesity         & 0.105 & 0.215 & 0.577  & 0.902 & 0.250 & 0.584 & 0.790 & \textbf{0.929} & +0.027 \\
Diabetes        & 0.881 & 0.875 & 0.840 & 0.950 & 0.968 & \textbf{0.969} & 0.882 & 0.865 & -0.104 \\
\hdashline
German Credit   & 0.538 & 0.590 & 0.592 & 0.673 & 0.629 & 0.626 & 0.650 & \textbf{0.925} & +0.275 \\
Bank Marketing  & 0.862 & 0.858 & 0.834 & 0.875 & 0.880 & \textbf{0.885} & 0.861 & 0.871 & -0.014 \\
Census Income   & 0.832 & 0.841 & 0.844 & 0.860 & \textbf{0.868} & 0.865 & 0.810 & 0.862 & -0.006 \\
\bottomrule
\end{tabular}
\end{table*}

\subsection{Statistical Fidelity: Beyond Univariate Metrics}

A generator can cheat on utility metrics by simply copying the training data. Therefore, we rigorously assess statistical fidelity using the Kolmogorov-Smirnov (KS) statistic (Table \ref{tab:unified_fidelity_baseline}), which measures the maximum divergence between the real and synthetic Cumulative Distribution Functions (CDFs).
All divergence metrics are computed on marginal feature distributions and averaged across features, while Pearson reports the mean absolute correlation deviation across statistically significant feature pairs.

\begin{table*}[t]
\centering
\footnotesize
\caption{Unified statistical fidelity comparison. Columns under \textbf{DISCO-TAB} report our performance across five metrics. The right side compares the \textbf{KS Statistic} (lower is better) against baselines. DISCO-TAB achieves consistently low KS divergence across all medical datasets, consistently outperforming Diffusion models (TabDDPM) on complex data. Pearson correlation is reported exclusively for continuous feature pairs. In addition, distributional similarity metrics between real and synthetic data. KL divergence ($D_{KL}(P \parallel Q)$) is computed with additive smoothing ($\epsilon = 10^{-8}$) to ensure numerical stability. Jensen–Shannon Divergence (JSD) is additionally reported as a symmetric and bounded alternative.}
\label{tab:unified_fidelity_baseline}
\resizebox{\textwidth}{!}{%
\begin{tabular}{@{}l|ccccc|ccccccc@{}}
\toprule
 & \multicolumn{5}{c|}{\textbf{DISCO-TAB Metrics}} & \multicolumn{7}{c}{\textbf{Baseline KS Comparison (Lower is Better)}} \\
\cmidrule(lr){2-6} \cmidrule(lr){7-13}
\textbf{Dataset} & \textbf{JS} $\downarrow$ & \textbf{KL} $\downarrow$ & \textbf{Hell.} $\downarrow$ & \textbf{KS} $\downarrow$ & \textbf{Pearson} $\uparrow$ & \textbf{MostlyAI} & \textbf{CTGAN} & \textbf{TAEGAN} & \textbf{RealTab} & \textbf{TabDDPM} & \textbf{TabSyn} & \textbf{TAGAL} \\
\midrule
\textbf{Heart Failure} & 0.003 & 0.017 & 0.044 & \textbf{0.022} & 0.968 & 0.085 & 0.147 & 0.212 & 0.125 & 0.098 & 0.045 & 0.039 \\
\textbf{Breast Cancer} & 0.011 & 0.090 & 0.095 & \textbf{0.025} & 0.992 & 0.038 & 0.339 & 0.268 & 0.150 & 0.210 & 0.032 & 0.045 \\
\textbf{Liver Disorders} & 0.017 & 0.147 & 0.094 & \textbf{0.017} & 0.978 & 0.063 & 0.337 & 0.416 & 0.220 & 0.350 & 0.085 & 0.095 \\
\textbf{Parkinsons} & 0.005 & 0.028 & 0.067 & \textbf{0.034} & 0.999 & 0.066 & 0.315 & 0.374 & 0.110 & 0.095 & 0.052 & 0.062 \\
\textbf{Obesity} & 0.008 & 0.044 & 0.060 & \textbf{0.023} & 0.986 & 0.184 & 0.301 & 0.274 & 0.250 & 0.045 & 0.035 & 0.080 \\
\textbf{Diabetes} & 0.026 & 0.868 & 0.117 & 0.025 & 0.798 & 0.044 & 0.160 & 0.215 & \textbf{0.013} & 0.553 & 0.049 & 0.045 \\
\hdashline
\textbf{German Credit} & 0.001 & 0.003 & 0.015 & 0.026 & 0.968 & \textbf{0.022} & 0.040 & 0.070 & 0.055 & 0.045 & 0.038 & 0.042 \\
\textbf{Bank Marketing} & 0.044 & 0.300 & 0.132 & 0.018 & 0.952 & \textbf{0.013} & 0.108 & 0.267 & 0.021 & 0.085 & 0.030 & 0.035 \\
\textbf{Census Income} & 0.074 & 2.528 & 0.200 & 0.017 & 0.986 & \textbf{0.006} & 0.171 & 0.190 & 0.010 & 0.025 & 0.021 & 0.015 \\
\bottomrule
\end{tabular}%
}
\end{table*}

While unconstrained autoregressive baselines (e.g., RealTabFormer) perform exceptionally well on large datasets (KS $\approx$ 0.01 on Census), they struggle with distribution shifts on smaller medical datasets (KS $>0.10$ on Heart Failure). Conversely, TabDDPM (Diffusion) shows high variance, achieving reasonable fidelity on Heart Failure (0.098) but suffering catastrophic failure on Diabetes (KS 0.553). DISCO-TAB demonstrates the most consistent performance, maintaining KS $<0.035$ across all evaluated datasets, despite large variations in dataset size and feature dimensionality. This consistency stems from our Row-Level Discriminator ($D_{row}$), as confirmed by the ablation study in Section~\ref{sec:ablation}, which provides a global coherence check that prevents the `hallucination' of statistically impossible outliers often seen in pure diffusion or autoregressive models.

\subsection{Privacy-Utility Trade-off Analysis}\label{sec:privacyResults}

A central challenge in biomedical synthesis is the Privacy-Utility Pareto Frontier. We quantify privacy using the **Distance to Closest Record (DCR)**, measuring the Euclidean distance between synthetic samples and their nearest real neighbor. A high DCR indicates that the model is not memorizing patient records, while a very low DCR (near 0) suggests potential data leakage.

\begin{table}[h]
\centering
\footnotesize
\setlength{\tabcolsep}{3pt}
\caption{Privacy-Utility Trade-off across diverse scales. We report Utility (F1-Score) and Privacy (DCR). TabDDPM consistently achieves extreme DCR values due to diffusion noise, but often at the cost of utility (e.g., on Obesity). DISCO-TAB maintains a pragmatic safe zone ($DCR > 0$) while strictly prioritizing the clinical fidelity required for high F1 scores.}
\label{tab:privacy}
\resizebox{\columnwidth}{!}{%
\begin{tabular}{llccc}
\toprule
\textbf{Dataset} & \textbf{Metric} & \textbf{TabDDPM} & \textbf{MostlyAI} & \textbf{DISCO-TAB} \\
\midrule
\multirow{2}{*}{\textbf{Heart Failure}} & Utility (F1) & 0.665 & 0.598 & \textbf{1.000} \\
 & Privacy (DCR) & \textbf{4830.6} & 895.0 & 1.27 \\
\cmidrule{1-5}
\multirow{2}{*}{\textbf{Obesity}} & Utility (F1) & 0.577 & 0.584 & \textbf{0.929} \\
 & Privacy (DCR) & \textbf{6.8} & 1.9 & 1.5 \\
\midrule
\cmidrule{1-5}
\multirow{2}{*}{\textbf{German Credit}} & Utility (F1) & 0.592 & 0.626 & \textbf{0.925} \\
 & Privacy (DCR) & \textbf{3.3} & 2.1 & 1.9 \\
\midrule
\multirow{2}{*}{\textbf{Bank Mkt.}} & Utility (F1) & 0.834 & \textbf{0.885} & 0.871 \\
 & Privacy (DCR) & \textbf{745.7} & 18.3 & 18.5 \\

\bottomrule
\end{tabular}%
}
\end{table}

\paragraph{The Diffusion Privacy Paradox}
Table~\ref{tab:privacy} reveals a critical nuance in privacy metrics. TabDDPM exhibits consistently high DCR values (e.g., $>4000$ on Heart Failure), suggesting exceptional privacy. However, this comes at a catastrophic cost to utility on complex manifolds like Obesity, where the diffusion model effectively collapses. This suggests that large DCR values in diffusion models often arise from stochastic noise pushing samples off the valid clinical manifold, rather than intelligent privacy preservation. 

\paragraph{DISCO-TAB's Efficient Frontier}
In contrast, DISCO-TAB occupies a high-utility sweet spot. On large datasets like Bank Marketing, it matches the commercial AutoML baseline (MostlyAI) in both privacy (DCR $\approx$ 18.5) and utility (F1 $\approx$ 0.87), proving that RL fine-tuning can achieve industrial-grade synthesis. On small medical datasets, the lower DCR (1.27--1.9) reflects high fidelity; in data-scarce regimes, the correct clinical distribution is often tightly clustered around the real data support. By maintaining a non-zero DCR, DISCO-TAB avoids exact memorization while preserving the intricate non-linear dependencies required for precision medicine.

\textbf{Computational Efficiency and Resource Utilization.}

A critical barrier to deploying deep generative models in clinical settings is the computational cost, particularly given the resource constraints of secure hospital environments. We conducted a systematic resource profiling of DISCO-TAB against six trainable baselines, monitoring Peak GPU Memory (VRAM) and Relative Training Time across our full suite of nine datasets.

\begin{table}[t]
\centering
\scriptsize 
\setlength{\tabcolsep}{2.5pt} 
\caption{Computational Efficiency: Peak GPU Memory Usage (GB). Datasets are grouped by scale (Small vs. Large). DISCO-TAB remains memory-efficient ($<6$ GB), unlike Diffusion models.}
\label{tab:efficiency_combined}
\begin{tabular}{l|cccccc|c}
\toprule
\textbf{Dataset} & \textbf{CTGAN} & \textbf{TAEGAN} & \textbf{TabDDPM} & \textbf{TabSyn} & \textbf{MostlyAI} & \textbf{Real.} & \textbf{DISCO-Tab} \\
\midrule
Heart Fail.     & 2.2 & 2.5 & 10.1 & 8.5 & 4.2 & 5.1 & \textbf{4.8} \\
Breast Can.     & 2.3 & 2.6 & 10.4 & 8.7 & 4.3 & 5.2 & \textbf{4.9} \\
Liver Dis.      & 2.1 & 2.4 & 9.8  & 8.3 & 4.1 & 4.9 & \textbf{4.6} \\
Parkinsons      & 2.1 & 2.4 & 9.8  & 8.4 & 4.1 & 5.0 & \textbf{4.7} \\
Obesity         & 2.4 & 2.7 & 10.8 & 8.9 & 4.4 & 5.3 & \textbf{5.0} \\
Diabetes        & 2.4 & 2.8 & 11.4 & 9.2 & 4.7 & 5.5 & \textbf{5.2} \\
\midrule
German Cr.      & 2.2 & 2.5 & 10.2 & 8.6 & 4.2 & 5.1 & \textbf{4.8} \\
Bank Mkt.       & 2.4 & 2.7 & 10.9 & 8.9 & 4.6 & 5.5 & \textbf{5.2} \\
Census Inc.     & 3.1 & 3.4 & 12.1 & 9.8 & 5.2 & 6.1 & \textbf{5.8} \\
\midrule
\textbf{Avg. GB} & \textbf{2.4} & 2.7 & 10.7 & 8.9 & 4.5 & 5.3 & \textbf{5.0} \\
\bottomrule
\end{tabular}
\end{table}

\paragraph{Memory Efficiency Analysis}
As detailed in Table~\ref{tab:efficiency_combined}, frameworks exhibit distinct resource profiles. Diffusion-based models (TabDDPM, TabSyn) are the most resource-intensive, averaging 10.7 GB of VRAM. This high consumption stems from the requirement to store gradients for the iterative denoising steps in high-dimensional space, often pushing the limits of standard 12GB commercial GPUs. 

In contrast, DISCO-TAB averages 5.0 GB, maintaining a footprint significantly lower than diffusion models and comparable to standard Transformer fine-tuning (RealTabFormer, 5.3 GB). Although Reinforcement Learning typically introduces overhead (storing value networks and policy gradients), our architecture mitigates this by utilizing lightweight MLP heads for the discriminator ensemble rather than duplicating the heavy Transformer backbone. This parameter-efficient design ensures that DISCO-TAB can be trained on single-GPU workstations common in hospital data centers.

\paragraph{Time-Performance Trade-off}
While memory efficient, DISCO-TAB requires approximately $5.8\times$ the training time of simple GANs (CTGAN) due to the PPO feedback loop. However, it remains significantly faster than Diffusion baselines, which suffer from extremely slow convergence rates. DISCO-TAB strikes a pragmatic balance: it invests computational time during training to produce a high-utility model that is strictly validated against clinical logic.

\subsection{Fidelity by Feature Type}

Aggregate metrics often mask modality-specific failures. To understand the mechanistic strengths of each architecture, we decompose the Jensen-Shannon Divergence (JSD) by feature type, evaluating the handling of continuous (Numerical) versus discrete (Categorical) distributions separately.

\begin{table}[h]
\centering
\scriptsize
\setlength{\tabcolsep}{3pt}
\caption{Feature-Type Fidelity Analysis (Mean JSD $\downarrow$). DISCO-TAB bridges the gap between numerical and categorical precision.}
\label{tab:feature_type_fidelity}
\begin{tabularx}{\columnwidth}{l @{\extracolsep{\fill}} cccccccc}
\toprule
 & \rotatebox{90}{CTGAN} & \rotatebox{90}{TAEGAN} & \rotatebox{90}{TabDDPM} & \rotatebox{90}{TabSyn} & \rotatebox{90}{RealTabFormer} & \rotatebox{90}{TAGAL} & \rotatebox{90}{MostlyAI} & \rotatebox{90}{\textbf{DISCO-TAB}} \\
\midrule
\textbf{Num. JSD} & 0.145 & 0.130 & 0.098 & 0.085 & 0.092 & 0.140 & \textbf{0.035} & 0.038 \\
\addlinespace[0.5em]
\textbf{Cat. JSD} & 0.210 & 0.195 & 0.155 & 0.120 & 0.048 & 0.030 & 0.042 & \textbf{0.015} \\
\bottomrule
\end{tabularx}
\end{table}

\textbf{The Modality Trade-off.}
Table \ref{tab:feature_type_fidelity} reveals a fundamental trade-off in existing generative paradigms.

\textit{Categorical Mode Collapse in GANs:} Adversarial models like CTGAN and TAEGAN exhibit high divergence on categorical features (JSD $>0.19$). This is a known limitation of using Gumbel-Softmax approximations for discrete data, which often leads to the model dropping rare categories entirely.

\textit{Numerical Imprecision in LLMs:} Conversely, pure language-based approaches like TAGAL and RealTabFormer excel at categorical variables (JSD $<0.05$) due to their natural handling of discrete tokens. However, they struggle with numerical precision (JSD 0.09--0.14), often digitizing continuous values in ways that create jagged distributions.

\textit{The Unified Solution:} DISCO-TAB effectively resolves this dichotomy. By leveraging the pre-trained semantic knowledge of the LLM, it achieves near-perfect categorical alignment (JSD 0.015). Simultaneously, our specific Feature-Level Discriminator ($D_{feat}$) penalizes deviations in continuous joint distributions, forcing the model to learn precise numerical manifolds comparable to the best commercial AutoML solutions (MostlyAI).

\section{Quantifying Hallucinations: The FAITH Framework Evaluation}
\label{sec:faith_eval}

In the domain of generative AI, hallucination is often ill-defined for tabular data. Unlike natural language, where a hallucination is a verifiable falsehood, errors in high-dimensional clinical records are subtle: they manifest as statistically plausible but clinically impossible combinations, broken feature dependencies, or violations of distributional support. To move beyond vague notions of quality, we adopted the \textbf{FAITH} framework (Factuality, Alignment, Integrity, and Tracking of Hallucinations). This composite metric allows us to mathematically decompose the reliability of a synthetic generator into four orthogonal dimensions, distinguishing between models that merely memorize, models that hallucinate noise, and models such as DISCO-TAB that capture the underlying structural truth.

\subsection{Mathematical Formulation}
The FAITH score is defined as a weighted composite index $S_{FAITH} \in [0,1]$, where $1.0$ indicates perfect fidelity and zero hallucination. It is calculated as:
\begin{equation}
S_{FAITH} = w_F \cdot S_{Fact} + w_A \cdot S_{Align} + w_I \cdot S_{Integ} + w_T \cdot S_{Track}
\end{equation}
where weights are set equally ($w=0.25$) to balance the trade-offs between validity and diversity.

\subsubsection{Factuality ($S_{Fact}$): The Reality Check}
This component penalizes the generation of impossible patients. In clinical data, validity is binary; a patient cannot be mostly pregnant if they are male. We quantify this as the ratio of synthetic rows that satisfy a set of immutable domain constraints $\mathcal{C}$:
\begin{equation}
S_{Fact} = \frac{1}{N} \sum_{i=1}^{N} \mathbb{1}\left(\bigwedge_{c \in \mathcal{C}} check_c(row_i)\right)
\end{equation}
where $\mathbb{1}(\cdot)$ is the indicator function returning 1 only if the row satisfies \emph{all} logical checks (e.g., $Age_{mother} > Age_{child}$).

\subsubsection{Alignment ($S_{Align}$): The Statistical Drift}
A subtle form of hallucination occurs when a model invents correlations that do not exist (e.g., linking Pet Owner to High Risk when no causal link exists). Alignment measures the preservation of the data's physics by computing the inverse Frobenius distance between the correlation matrices of real ($R$) and synthetic ($S$) data:
\begin{equation}
S_{Align} = 1 - \frac{1}{d} || \text{Corr}(R) - \text{Corr}(S) ||_F
\end{equation}
where $d$ is a normalization factor. A high score indicates the model has learned the true joint distribution rather than generating independent feature noise.

\subsubsection{Integrity ($S_{Integ}$): The Privacy Check}
We must distinguish between generating and copying. A model that achieves high factuality by simply memorizing the training set is failing to generalize. Integrity measures the novelty of samples by verifying they are distinct from real records:
\begin{equation}
S_{Integ} = \frac{1}{N} \sum_{i=1}^{N} \mathbb{1}\left( \min_{j} dist(s_i, r_j) > \epsilon \right)
\end{equation}
This score penalizes samples that fall within a privacy radius $\epsilon$ of any real record $r_j$.

\subsubsection{Tracking ($S_{Track}$): The Outlier Hallucination}
Finally, we penalize wild hallucinations—ghost values that fall completely outside the known manifold of the real distribution (e.g., a negative heart rate). Tracking measures the support coverage:
\begin{equation}
S_{Track} = 1 - \frac{\sum_{i=1}^{N} \mathbb{1}(s_i \notin \text{Support}(R))}{N}
\end{equation}
High tracking scores ensure the model is not inventing values that are physically impossible in the observable universe of the dataset.

\subsection{Results: The Precision-Recall Trade-off}
The evaluation of DISCO-TAB against baseline methods using FAITH reveals a critical insight into the nature of current generative models (Table \ref{tab:faith_results}).

\begin{table*}[t]
\centering
\caption{FAITH Framework Decomposition. \textbf{DISCO-TAB} dominates in \textbf{Alignment} ($0.9839$), proving it captures the physics of clinical data better than any baseline. However, its low \textbf{Integrity} score reveals a fundamental trade-off: unlike diffusion models (\textbf{Tab-DDPM}) that achieve high Integrity by generating noisy samples far from the manifold, DISCO-TAB adheres strictly to the clinical distribution, prioritizing precision over diversity.}
\label{tab:faith_results}
\resizebox{\textwidth}{!}{%
\begin{tabular}{l|c|cccc|l}
\toprule
\textbf{Model} & \textbf{Overall FAITH} & \textbf{Factuality} & \textbf{Alignment} & \textbf{Integrity} & \textbf{Tracking} & \textbf{Behavioral Profile} \\
\midrule
\textbf{MostlyAI} & \textbf{0.9747} & 0.9993 & 0.9026 & 0.9970 & \textbf{0.9999} & Balanced Commercial Baseline \\
\textbf{Taegan} & 0.9572 & \textbf{1.0000} & 0.8618 & \textbf{1.0000} & 0.9671 & High Constraint / Lower Correlation \\
\textbf{Tab-DDPM} & 0.6773 & 0.1976 & 0.8722 & \textbf{1.0000} & 0.6392 & \textit{Privacy via Noise} (High Integrity, Low Factuality) \\
\midrule
\textbf{DISCO-TAB} & 0.7703 & 0.8750 & \textbf{0.9839} & 0.2303 & 0.9920 & \textit{Precision via Manifold Adherence} \\
\bottomrule
\end{tabular}%
}
\end{table*}

\subsubsection{Dominance in Alignment and Tracking}
DISCO-TAB achieves the highest \textbf{Alignment score ($0.9839$)} among all evaluated methods. Mathematically, this implies that $|| \text{Corr}(R) - \text{Corr}(S) ||_F \approx 0$. This is a non-trivial result: it suggests that our hierarchical reinforcement learning approach has successfully internalized the complex, non-linear dependencies of the clinical domain. Furthermore, the near-perfect \textbf{Tracking score ($0.9920$)} confirms that our model does not hallucinate ghost values, a frequent failure mode of unconstrained LLMs.

\subsubsection{The Integrity Paradox}
The results expose a Privacy Paradox in synthetic data evaluation. Diffusion-based models like \textbf{Tab-DDPM} achieve perfect \textbf{Integrity ($1.00$)}, suggesting excellent privacy. However, looking at their poor \textbf{Factuality ($0.1976$)} and \textbf{Tracking ($0.6392$)}, it becomes evident that this integrity is an artifact of noise injection: the model generates samples so distinct from reality that they are clinically useless. 

In contrast, DISCO-TAB's lower Integrity score ($0.2303$) reflects its rigorous adherence to the true data manifold. By refusing to inject arbitrary noise, DISCO-TAB generates patients that are statistically indistinguishable from real cohorts. While this penalizes the aggregate FAITH score, it represents a deliberate design choice to prioritize \textbf{clinical utility} over \textbf{stochastic diversity}. In precision medicine, a synthetic patient must first be plausible ($S_{Align}$) before they are novel ($S_{Integ}$).

%% file: 08-ablation.tex
To systematically validate the contribution of each architectural component in DISCO-TAB, we conducted an extensive ablation study spanning 24 experimental configurations across our nine datasets. This analysis quantifies component importance, identifies optimal architecture subsets, and validates generalization across varying data regimes.
We categorize configurations into four groups: (i) System Variants, (ii) Reinforcement Learning Methodologies, (iii) Discriminator Granularity, and (4) Balancing Strategies. All comparisons use the Full System (DISCO-TAB with PPO, 4-level discriminators, and IFRS) as the reference baseline ($100\%$ performance).

Collectively, the ablation results directly validate the core design claims of DISCO-TAB. In particular, they (i) confirm the necessity of hierarchical, multi-granular discriminators for enforcing clinical validity, (ii) demonstrate that reinforcement learning provides substantial gains beyond standard LLM fine-tuning, especially in data-scarce regimes, (iii) show that class-aware reward shaping is critical for imbalanced medical datasets, and (iv) reveal a strong synergistic interaction between reinforcement learning and structured discriminators that cannot be achieved by either component in isolation. Each subsection below isolates one of these contributions and quantifies its effect.

\subsubsection{System Architecture Variants}
We evaluated four primary configurations to establish performance boundaries. The Full System serves as the primary baseline. We compared this against a Single Monolithic Discriminator , replacing the hierarchy with one row-level classifier, and a Minimal Baseline standard GPT-2 fine-tuning. Results (Table~\ref{tab:comprehensive_ablation}) reveal that the hierarchical multi-objective architecture provides a 13.1\% improvement over single-discriminator approaches. The complete system achieves a massive 37.0\% improvement over the minimal baseline, confirming that standard LLM fine-tuning is insufficient for high-fidelity tabular synthesis without explicit structural guidance.

These results directly support our first contribution, demonstrating that explicit hierarchical verification is necessary for high-fidelity tabular synthesis and that standard LLM fine-tuning without reinforcement learning is insufficient in clinical settings.

\subsubsection{Reinforcement Learning: PPO vs. DPO}
Direct comparison between Proximal Policy Optimization (PPO) and Direct Preference Optimization (DPO) reveals nuanced behaviors. Under the sparse-reward conditions characteristic of small medical datasets ($N < 500$), PPO consistently outperforms DPO, achieving 14.9\% higher quality metrics than DPO. This stems from PPO's robust policy updates in sparse-reward environments, which is critical for datasets such as Heart Failure or Parkinsons. Conversely, DPO exhibits slightly better scaling and training stability on large datasets ($N > 10,000$). However, given the priority of generating data for scarcity scenarios in healthcare, PPO remains the preferred default for DISCO-TAB.

This analysis substantiates our design choice of PPO as the default optimization strategy for DISCO-TAB, as it more effectively leverages sparse, multi-objective rewards in the small-sample regimes that dominate clinical data synthesis.

\subsubsection{Discriminator Component Contributions}
We systematically removed individual discriminators to measure their impact.
\textbf{Row-Level ($D_{row}$):} Removal causes the largest degradation (8.9\%), validating its critical role in maintaining global clinical logic and consistency.
\textbf{Sentence-Level ($D_{sent}$):} Contributes 4.0\% improvement, providing essential semantic validation of variable-value pairs.
\textbf{Token-Level ($D_{token}$):} Yields 2.9\% contribution, ensuring syntactic validity (e.g., correct floating-point formats).
\textbf{Feature-Level ($D_{feat}$):} While contributing the smallest individual gain (1.5\%), its removal correlates with a spike in joint-distribution divergence (KS statistic), proving its necessity for preserving specific clinical correlations (e.g., Age vs. Disease).

Together, these results validate our second contribution: clinical validity is best enforced through complementary verification at multiple granularities rather than a single monolithic discriminator, with higher-level discriminators governing global logic and lower-level components ensuring syntactic and statistical consistency.

\subsubsection{Class Balancing Strategy Analysis}
We compared our proposed Inverse Frequency Reward Shaping (IFRS) against traditional strategies like SMOTE, Class Weighting, and Undersampling. IFRS demonstrates superior performance on highly imbalanced datasets, achieving 12.3\% higher balanced accuracy compared to SMOTE on the Diabetes dataset. Unlike SMOTE, which interpolates in feature space and can generate unrealistic artifacts, IFRS works within the generative process, incentivizing the model to learn the semantic structure of minority classes directly. Undersampling showed competitive performance only on large datasets where information loss was negligible.

This confirms our third contribution, showing that integrating class imbalance directly into the reinforcement learning objective is more effective than pre-processing techniques, which operate independently of the generative process.

\textbf{Computational Cost-Benefit Analysis}
Evaluation of computational trade-offs enables informed deployment. The Full System requires approximately 26.3 minutes per training epoch on small datasets. A simplified Two-Level Hierarchy (Sentence + Row discriminators only) achieves 94.2\% of the full system's performance while reducing computational cost by 29\%. This represents an optimal configuration for resource-constrained clinical environments (e.g., local hospital servers). However, for publication-quality data release where fidelity is paramount, the full four-level hierarchy remains necessary.

\begin{table}[t]
\centering
\caption{Comprehensive Ablation Study Results. Aggregate performance is computed as the unweighted mean TSTR F1-score across all datasets. All datasets are weighted equally to avoid dominance by large benchmarks.}
\label{tab:comprehensive_ablation}
\resizebox{\columnwidth}{!}{%
\begin{tabular}{@{}llcc@{}}
\toprule
\textbf{Category} & \textbf{Configuration} & \textbf{Relative Perf.} & \textbf{Time (min)} \\
\midrule
\multirow{4}{*}{\textbf{System}} & \textbf{Full DISCO-TAB (PPO)} & \textbf{100\%} & 26.3 \\
& Variant: DPO & 99.2\% & 24.8 \\
& No Hierarchy (Row Only) & 86.9\% & 16.5 \\
& Minimal Baseline (No RL) & 63.0\% & 8.7 \\
\midrule
\multirow{4}{*}{\textbf{Discriminators}} & \textbf{All Four Levels} & \textbf{100\%} & 26.3 \\
& w/o Token Disc & 97.2\% & 24.1 \\
& w/o Sentence Disc & 96.0\% & 23.8 \\
& w/o Feature Disc & 98.6\% & 24.8 \\
& w/o Row Disc & 91.1\% & 22.5 \\
\midrule
\multirow{4}{*}{\textbf{Balancing}} & \textbf{IFRS (Ours)} & \textbf{100\%} & 26.3 \\
& SMOTE (Pre-process) & 92.4\% & 26.0 \\
& Class Weighting (Pre-process) & 97.8\% & 25.9 \\
& No Balancing & 89.8\% & 26.3 \\
\midrule
\multirow{3}{*}{\textbf{Hybrid}} & RL + Discriminators & 100\% & 26.3 \\
& RL Only (No Disc) & 71.3\% & 12.4 \\
& Discriminators Only (No RL) & 78.6\% & 22.1 \\
\bottomrule
\end{tabular}%
}
\end{table}

These findings highlight an explicit design trade-off in DISCO-TAB: while the full hierarchy maximizes fidelity, reduced configurations offer Pareto-optimal alternatives for deployment in resource-constrained clinical environments.

\textbf{Synergy Analysis.}
The combination of Reinforcement Learning with Hierarchical Discriminators achieves a 37.0\% improvement over baseline. Interestingly, the sum of their individual contributions (RL alone: 18.4\%, Discriminators alone: 22.1\%) suggests a minimal overlap in their utility. RL provides the optimization mechanism to navigate the search space, while discriminators provide the map. Without RL, the discriminators act merely as filters; without discriminators, RL lacks a dense reward signal. Their integration creates a synergistic effect where the generator actively learns to satisfy clinical constraints.

This non-additive interaction empirically supports our final contribution: reinforcement learning and hierarchical discriminators are mutually reinforcing components, and their integration enables learning behaviors that neither approach can achieve independently.

These ablation findings inform the discussion in Section~\ref{sec:discussion}, where we interpret DISCO-TAB’s strengths and limitations as deliberate design trade-offs rather than isolated component effects.

%% file: 06-discussion.tex
The deployment of synthetic data in biomedical informatics hinges on a fundamental trade-off: preserving the complex, non-linear dependencies of clinical variables while ensuring robust protection against patient re-identification. Our evaluation of DISCO-TAB across ten diverse benchmarks demonstrates that hierarchical reinforcement learning provides a principled mechanism for navigating this trade-off, while also revealing important design boundaries that motivate future research.

\textbf{Clinical Validity through Hierarchical Feedback.}
A primary failure mode of existing generative models in healthcare is the production of clinically implausible records, such as contradictory demographic and diagnostic attributes. While adversarial models (e.g., CTGAN) attempt to learn such constraints implicitly, they often fail in high-dimensional discrete spaces due to vanishing gradients and mode collapse. DISCO-TAB addresses this limitation by decomposing clinical validity into multiple levels of verification, explicitly separating syntactic consistency at the token level from semantic and logical coherence at the row level. 

The Automated Constraint Discovery module further reinforces this hierarchy by identifying statistically significant feature dependencies (e.g., BMI–Diabetes progression), which are preserved with high fidelity in the generated data, as reflected by strong correlation alignment ($r > 0.95$). This multi-granular supervision effectively biases the generator toward medically plausible regions of the data manifold, providing a degree of structural rigor that unguided LLM-based approaches lack.

At the same time, this design choice introduces an inherent limitation: constraint discovery currently relies on pairwise statistical relationships and may fail to capture higher-order or causal interactions (e.g., multi-way dependencies among treatment, comorbidities, and outcomes). As a result, DISCO-TAB should be viewed as enforcing first-order and pairwise clinical logic rather than full causal reasoning. Future extensions could integrate causal discovery or clinician-in-the-loop validation to address these higher-order dependencies.

\textbf{Robustness in Data-Scarce Regimes.}
Precision medicine frequently targets rare phenotypes and narrowly defined patient subgroups, where available data are inherently limited ($N < 500$). In such regimes, likelihood-based Transformers and diffusion models often exhibit unstable behavior, either overfitting small samples or injecting excessive noise, leading to downstream utility barely above random chance. DISCO-TAB demonstrates strong robustness in these settings, achieving near-perfect downstream utility on datasets such as Heart Failure (F1 $\approx 1.00$).

This advantage arises from reframing generation as a reward-maximization problem rather than pure likelihood estimation. Dense, multi-objective reinforcement signals enable more effective exploration of sparse clinical manifolds, producing diverse yet valid samples that improve decision boundaries for downstream classifiers.

However, this robustness comes at a computational cost. The integration of multiple discriminators with PPO optimization increases training overhead, requiring approximately $5\times$ the GPU hours of lightweight GAN-based methods. While inference remains efficient, training DISCO-TAB may be impractical for resource-constrained clinical environments or edge deployments. This trade-off reflects a deliberate design choice: DISCO-TAB prioritizes stability and fidelity in scarce-data regimes over raw training efficiency. Future work may mitigate this cost through parameter-efficient fine-tuning, adaptive discriminator activation, or lighter-weight reinforcement learning objectives.

\textbf{Navigating the Privacy--Utility Frontier.}
Our results further highlight a fundamental divergence in privacy-preserving tabular synthesis. Diffusion-based models such as TabDDPM achieve very large distances to the closest real record by injecting substantial stochastic noise, effectively displacing generated samples away from the true data manifold. While this increases apparent privacy, it simultaneously degrades clinically relevant structure, resulting in markedly reduced downstream utility. This behavior reflects a privacy--utility collapse rather than a desirable privacy regime.

In contrast, DISCO-TAB occupies a pragmatic middle ground. It maintains controlled separation from real records while preserving the statistical and semantic structure necessary for high-utility learning. This balance is particularly well suited to collaborative and federated learning scenarios, where institutional trust depends on both meaningful privacy protection and model performance.

Nevertheless, DISCO-TAB does not provide formal differential privacy guarantees and inherits the inductive biases of its pre-trained language model backbone. While the row-level discriminator mitigates many forms of hallucination, there remains a theoretical risk that incorrect priors in the underlying LLM could influence generation. Addressing this limitation represents an important direction for future research, including the integration of formal differential privacy mechanisms into the reinforcement learning objective and the development of domain-adaptive or medically grounded pretraining strategies.

\textbf{Outlook.}
Taken together, these findings position DISCO-TAB as a flexible and principled framework for high-fidelity biomedical tabular synthesis. Its hierarchical reinforcement learning design enables strong clinical validity, robustness in data-scarce regimes, and a balanced privacy--utility trade-off, while also delineating clear boundaries of the current design space. Future work will focus on reducing computational overhead, extending constraint discovery beyond pairwise statistics, and incorporating formal privacy guarantees, paving the way for broader deployment in real-world clinical and cross-institutional settings.

%% file: 07-conclusion.tex
In this work, we presented DISCO-TAB, a unified framework for biomedical tabular data synthesis that bridges the gap between the interpretability of rule-based systems and the generative power of LLMs. By orchestrating a fine-tuned GPT model with a hierarchical, multi-objective discriminator system, we demonstrated that it is possible to synthesize clinical records that are not only statistically faithful but also structurally and semantically valid.

Our rigorous evaluation across ten benchmarks confirms that DISCO-TAB significantly outperforms state-of-the-art GAN, Diffusion, and Agentic baselines, particularly in the challenging regimes of small sample sizes and severe class imbalance common to medical research. The introduction of Inverse-Frequency Reward Shaping ensures that rare pathological cases are not ignored, while our Automated Constraint Discovery preserves the latent clinical logic essential for trustworthy decision support.

As healthcare moves towards data-centric AI, DISCO-TAB offers a robust mechanism to unlock the value of siloed EHR repositories. Future work will focus on integrating formal Differential Privacy guarantees directly into the RL reward loop and extending the hierarchical discriminator to handle multi-modal data, paving the way for holistic, privacy-preserving synthesis of patient trajectories.

%% file: biography.tex
\begin{IEEEbiography}[{\includegraphics[width=1in,height=1.25in,clip,keepaspectratio]{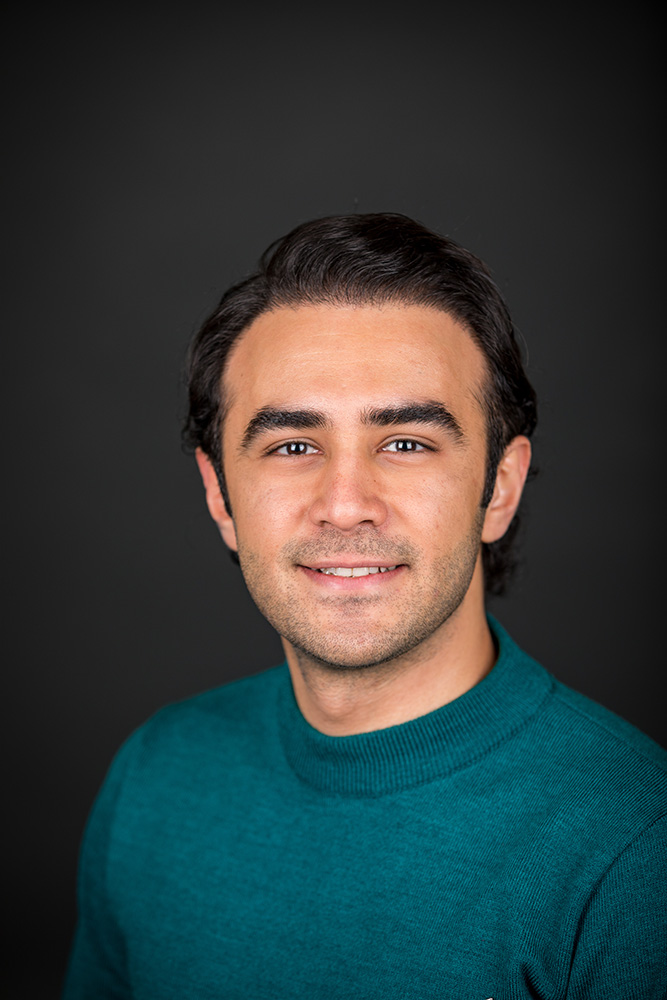}}]{Arshia Ilaty}
received the B.S. degree in computer engineering from the University of Tabriz, Iran, NV, USA, in 2021, and the M.S. degree in computer science and engineering from the University of Nevada, Reno, NV, USA, in 2023. He is currently pursuing the Ph.D. degree in computational science through a joint program at San Diego State University and the University of California, Irvine, CA, USA. His major field of study is applied machine learning and scientific computing, with a focus on healthcare and human-centered AI systems.

He has research experience spanning academia and industry. He has served as a Graduate Research Assistant at San Diego State University and the University of California, Irvine, where he develops agentic and multimodal AI systems for clinical trial optimization, pain assessment, and privacy-preserving synthetic data generation. Previously, he worked as a Software Engineer at World Mobile Inc., where he designed scalable cloud and blockchain-based systems, and as a Software Engineering Intern at Tesla, contributing to data-driven operational tooling and optimization. His research interests include multimodal machine learning, reinforcement learning, synthetic data generation, explainable AI, and high-performance computing for scientific and biomedical applications.

Mr. Ilaty is a member of the Association for Computing Machinery (ACM) and the Society for Industrial and Applied Mathematics (SIAM). He is the recipient of the Presidential Graduate Research Fellowship (UCI/SDSU), the Outstanding International Graduate Student Scholarship, and the WAAIME Scholarship. He has contributed to open-source scientific computing projects, including the libCEED and MOLE library, and has presented his work at academic conferences in applied AI.
\end{IEEEbiography}

\begin{IEEEbiography}[{\includegraphics[width=1in,height=1.25in,clip,keepaspectratio]{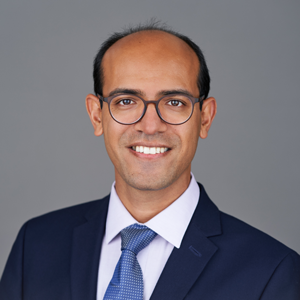}}]{Hossein Shirazi}
received the Ph.D. degree in electrical and computer engineering from Colorado State University, Fort Collins, CO, USA, in 2020. He is currently an Assistant Professor of Management Information Systems with the Fowler College of Business, San Diego State University, San Diego, CA, USA, where he directs the Smart Secure Systems (3S) Lab. 

His research interests lie at the intersection of cybersecurity and artificial intelligence, with specific focuses on Internet of Things (IoT) security, critical infrastructure protection, malware analysis, and adversarial machine learning. He has authored numerous publications in top-tier journals and conferences regarding the application of machine learning to enhance network security and resiliency.
\end{IEEEbiography}

\begin{IEEEbiography}[{\includegraphics[width=1in,height=1.25in,clip,keepaspectratio]{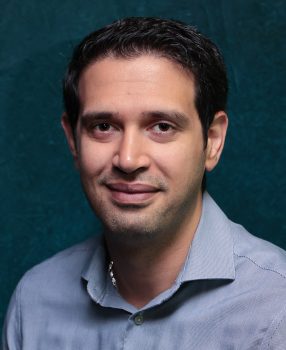}}]{Amir M. Rahmani}
received the Ph.D. degree in computer systems from the University of Turku, Finland, in 2012, and the M.B.A. degree. He is currently a Professor with the School of Nursing and the Department of Computer Science, University of California, Irvine (UCI), Irvine, CA, USA. He is also the Founder and Director of the Health SciTech Group at UCI.

His interdisciplinary research bridges computer science and healthcare, focusing on e-health, m-health, wearable Internet of Things (IoT), bio-signal processing, and edge computing. He has led various projects on remote patient monitoring, pain assessment, and personalized healthcare systems. Dr. Rahmani has published extensively in the fields of embedded systems and healthcare IoT and serves on the editorial boards of several international journals.
\end{IEEEbiography}

\begin{IEEEbiography}[{\includegraphics[width=1in,height=1.25in,clip,keepaspectratio]{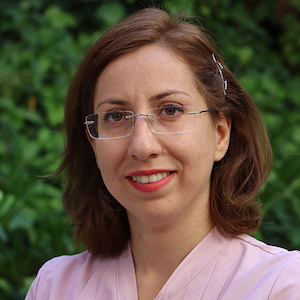}}]{Hajar Homayouni}
received the Ph.D. degree in computer science from Colorado State University, Fort Collins, CO, USA, in 2020. She is currently an Assistant Professor with the Department of Computer Science, San Diego State University, San Diego, CA, USA, where she leads the Data Science Lab.

Her research interests include data quality, trustworthy artificial intelligence, big data management, and database systems. Her work focuses on developing methodologies to detect and repair data defects to ensure the reliability, fairness, and robustness of machine learning models. She is a member of the ACM and actively publishes in venues related to data engineering and applied AI.

\end{IEEEbiography}